\pdfoutput=1

\documentclass[11pt]{article}

\usepackage[review]{acl}

\usepackage{times}
\usepackage{latexsym}

\usepackage[T1]{fontenc}

\usepackage[utf8]{inputenc}

\usepackage{microtype}

\usepackage{inconsolata}

\usepackage{graphicx}

\usepackage{algorithm}
\usepackage{algorithmic}
\usepackage{amsfonts}
\usepackage{amsmath}
\usepackage{tabularx}
\usepackage{multirow}   
\usepackage{booktabs}   
\usepackage{makecell}

\title{Adversarial Attacks on VQA-NLE: Exposing and Alleviating Inconsistencies in Visual Question Answering Explanations}

\author{Yahsin Yeh, Yilun Wu, Bokai Ruan, Honghan Shuai \\
National Yang Ming Chiao Tung University}

\begin{document}
\maketitle
\begin{abstract}
Natural language explanations in visual question answering (VQA-NLE) aim to make black-box models more transparent by elucidating their decision-making processes. However, we find that existing VQA-NLE systems can produce inconsistent explanations and reach conclusions without genuinely understanding the underlying context, exposing weaknesses in either their inference pipeline or explanation-generation mechanism. To highlight these vulnerabilities, we not only leverage an existing adversarial strategy to perturb questions but also propose a novel strategy that minimally alters images to induce contradictory or spurious outputs. We further introduce a mitigation method that leverages external knowledge to alleviate these inconsistencies, thereby bolstering model robustness. Extensive evaluations on two standard benchmarks and two widely used VQA-NLE models underscore the effectiveness of our attacks and the potential of knowledge-based defenses, ultimately revealing pressing security and reliability concerns in current VQA-NLE systems.

\end{abstract}

\section{Introduction}

Visual Question Answering with Natural Language Explanations (VQA-NLE) has recently become an active area of research \cite{nlxgpt, s3c, contrastivelearning}, as it promises both accurate answers and human-readable justifications. By augmenting conventional VQA pipelines with textual rationales, it can offer deeper transparency and facilitate trust in black-box models \cite{evil}. Moreover, generating explanations has been shown to reinforce question-answering performance itself, surpassing models trained solely on image-question pairs \cite{evil}. Despite this potential, critical questions remain about the quality and consistency of the explanations produced, prompting further investigation into how these models truly reason about visual and linguistic inputs.

Specifically, while VQA-NLE models can produce explanations for their decisions, we observe that they can yield contradictory or inconsistent outputs even when the input scenario remains essentially the same. For instance, consider an image that depicts a woman wearing goggles and skiing downhill. If a person asks, ``Why is the woman wearing goggles?'' the model answers, ``to protect eyes because the woman is wearing goggles to protect eyes,'' along with an explanation mentioning the goggles. However, rephrasing the prompt slightly to ``Why is the woman using goggles?'' causes the system to respond, ``to photograph because the woman is using a camera,'' thereby contradicting its previous statement. Such inconsistent responses raise doubts about whether these models truly ground their reasoning in the image-question pair or instead rely on superficial cues, thereby questioning the extent to which VQA-NLE models genuinely ``understand'' their inputs when generating explanations.

To explore whether these contradictory outputs reflect genuine weaknesses in VQA-NLE systems, we leverage an existing attack and also propose a new adversarial sample generation framework designed to uncover vulnerabilities across both textual and visual modalities. In particular, we systematically rephrase questions (while preserving semantic equivalence) or selectively remove objects from images—even those seemingly irrelevant to the query. These controlled yet minimal edits often cause the model to produce inconsistent explanations, revealing the model’s reliance on shallow patterns rather than robust visual-textual reasoning.\footnote{Although we primarily evaluate text-based and image-based manipulations separately, in principle they can also be combined to further stress-test a model’s consistency.}

In addition to exposing such vulnerabilities, we present an alleviation strategy based on integrating external knowledge into the question. Concretely, for each query, we use a language model to generate short, relevant knowledge statements (e.g., clarifying synonyms or describing contextual details). Appending these statements to the input helps the VQA-NLE model anchor its reasoning in genuine semantic understanding rather than superficial cues. Experimental results demonstrate that this knowledge-driven approach can reduce contradictory explanations, offering a practical pathway toward more reliable and transparent VQA-NLE systems.

In summary, our contributions are as follows:

\begin{itemize}
    \item To the best of our knowledge, this is the first adversarial framework specifically aimed at revealing potential vulnerabilities in VQA-NLE models, offering a systematic way to probe their security and consistency.
    \item Our attacks indeed degrade the semantic consistency of the VQA-NLE models on standard benchmarks (VQA-X and A-OKVQA), highlighting the models’ reliance on brittle cues.
    \item We propose a knowledge-based mitigation strategy that reduces inconsistencies introduced by adversarial textual, improving robustness in VQA-NLE.
\end{itemize}
\section{Related Work}

\subsection{VQA Explanations}
Natural language explanations (NLEs) for question-answering (QA) have garnered increasing attention, driven by findings that explicit rationales can bolster a model’s reasoning capabilities and interpretability \cite{esnli, multimodalexplanations, wt5, chainofthought}. In the context of visual question answering (VQA), these so-called VQA-NLE methods provide human-readable justifications alongside predicted answers \cite{multimodalexplanations, evil}. Approaches broadly split into post hoc (predict first, then explain) \cite{multimodalexplanations, faithfulexplanation, evil} and self-rationalizing methods, where answer prediction and explanation generation occur jointly \cite{nlxgpt, s3c}. Recent work has introduced contrastive objectives to further align explanations with visual evidence \cite{contrastivelearning}. Despite improvements in accuracy, concerns remain regarding how faithfully these explanations reflect genuine model reasoning, as opposed to exploiting spurious patterns \cite{sunnydark, towardscausalvqa}. Few studies have probed how small alterations in text or images can undermine explanation consistency, leaving open questions about the robustness of VQA-NLE outputs.

\subsection{VQA Robustness}
Meanwhile, VQA robustness research has largely focused on ensuring answer consistency under various input perturbations. For instance, \cite{sunnydark, cycleconsistency} investigate how changes in question phrasing affect predictions, whereas \cite{towardscausalvqa} examine the impact of altering semantic elements in images. Augmentation techniques have also been proposed to mitigate inconsistent or brittle answers \cite{towardscausalvqa, counterfactual}. However, most such work overlooks natural language explanations, and the few efforts targeting NLE consistency \cite{makeupyourmind, know} address primarily text-based variations.
In contrast, we adopt a broader perspective on VQA-NLE robustness by implementing and introducing adversarial attack frameworks that target both linguistic and visual inputs. We then propose a knowledge-based defense to bolster model reliability against these minimal yet strategically chosen perturbations. By assessing both answer correctness and explanation consistency, our work expands robustness research into interpretability-focused VQA systems.
\section{Method}

While robustness in linguistic variations and image semantics has been respectively studied in the fields of language modeling and VQA, it remains an underexplored area for VQA-NLE models. To this end, we structure our attack method into two approaches, each targeting a different aspect: \textbf{text-based} attack and \textbf{image-based} attack.

\subsection{Text-based Attack}
For the text-based approach, existing adversarial attack methods for discrete data, such as BERT-Attack~\cite{bertattack} and R\&R~\cite{randr}, are well-developed for generating adversarial text samples. Here, we directly employ BERT-Attack to generate text perturbations with synonym-based word substitution, aiming to fool models while maintaining grammatical and semantic coherence. By doing so, we expect to expose the model's reliance on superficial linguistic cues rather than genuine contextual understanding. Specifically, the attack highlights cases where the model associates certain words or phrases with specific answers and explanations, revealing a dependency on dataset biases rather than robust reasoning.

To implement this attack, we leverage a masked language model (MLM) as part of the candidate ranking mechanism. The core objective function aims to rank substitution sequences based on contextual fluency. Specifically, given a set of token-level substitution candidates generated via Byte Pair Encoding (BPE), the method constructs full-length sequences by exhaustively combining candidates across token positions. Each sequence is then evaluated by computing the token-wise cross-entropy loss between the MLM's predicted distribution and the actual substitute token IDs. These losses are aggregated and exponentiated to estimate each sequence’s perplexity, which serves as a proxy for fluency. By ranking sequences in ascending order of perplexity, the model promotes those that are both semantically coherent and grammatically well-formed.

Building upon this ranking mechanism, we then apply a filtering step to ensure semantic consistency between the adversarial and original inputs. Concretely, given the original question $Q$, we first use BERT-Attack to generate an adversarial candidates set containing $n$ samples, denoted by $\{Q^{\prime}_i\}_{i=1}^n$. Afterwards, we compute the semantic similarity between $Q^{\prime}_i$ and the original question $Q$ using a universal sentence encoder $U_s$~\cite{use}. If the cosine similarity between $Q^{\prime}_i$ and $Q$, denoted by $\gamma_i$, is lower than the predefined threshold $\sigma_s$, i.e., $\gamma_i=\cos(U_s(Q^{\prime}_i), U_s(Q))<\sigma_s$, then the candidate $Q^{\prime}_i$ is rejected. Otherwise, $Q^{\prime}_i$ is considered a valid adversarial example for the given image $I$. Following the previous work~\cite{bae}, we set $\sigma_s=0.8$.

\subsection{Image-based Attack}
To evaluate the robustness of VQA-NLE models against semantic changes in images, we propose a pipeline for editing images corresponding to each question. We hypothesize that any change influencing the model’s predictions can reveal its weak contextual understanding. For instance, in Figure~\ref{img_attack_sample} (left), for the question about the type of the event, removing the table should not influence the model's prediction. To guarantee that the question's context remains unchanged after the image modification, we must ensure that objects relevant to the question's context are preserved. 

Specifically, to generate an adversarial image without altering its overall meaning, we first identify any objects referenced in the question and answer, then limit our edits to regions unrelated to those objects. As the commonly-used VQA-NLE datasets, e.g., VQA-X~\cite{multimodalexplanations} and A-OKVQA~\cite{aokvqa} datasets, contain images sourced from MS-COCO~\cite{mscoco}, we consider the 80 predefined object classes and ground truth bounding boxes. To remove objects from the image, we utilize a diffusion-based inpainting approach~\cite{powerpaint}, ensuring that the edited image remains semantically coherent. Our approach for maintaining contextual consistency in image modifications comprises two steps: (1) vocabulary mapping and (2) object removal.

\paragraph{Vocabulary Mapping}
To determine whether an object can be removed, we first map all object references in the question, answer, and explanation to the 80 COCO categories. These categories are often referred to using various synonyms or subset terms in the QA and explanation space. For example, van, taxi, trunk, truck, and SUV all correspond to the category ``car,'' while table and desk refer to the category ``dining table.'' To prevent erroneous removals, we compile a comprehensive mapping of nouns, pronouns, and synonyms used in the QA and explanation vocabulary to the 80 COCO categories. Due to the space constraint, the complete list of the mapping table is available in the supplementary materials.

\paragraph{Object Removal}
Let $S_{\text{I}}$ represent the set of objects in the images (identified via COCO bounding boxes), $S_{\text{QA}}$ represent the set of objects in the QA pair, and $S_{\text{E}}$ represent the set of objects in the explanation. We define the set of candidate objects for removal as
\begin{equation}
S_{\text{candidate}}:=S_{\text{E}} \cap \{ S_{\text{I}} \setminus S_{\text{QA}}\}.
\end{equation}
We then select the most frequent object in $S_{\text{candidate}}$ as our target object, $S_{\text{target}}$. Our underlying assumption is that a robust model should continue to generate explanations that accurately reflect the modified image content and do not mislead.

\begin{figure*}[!ht]
    \centering
    \includegraphics[width=\textwidth]{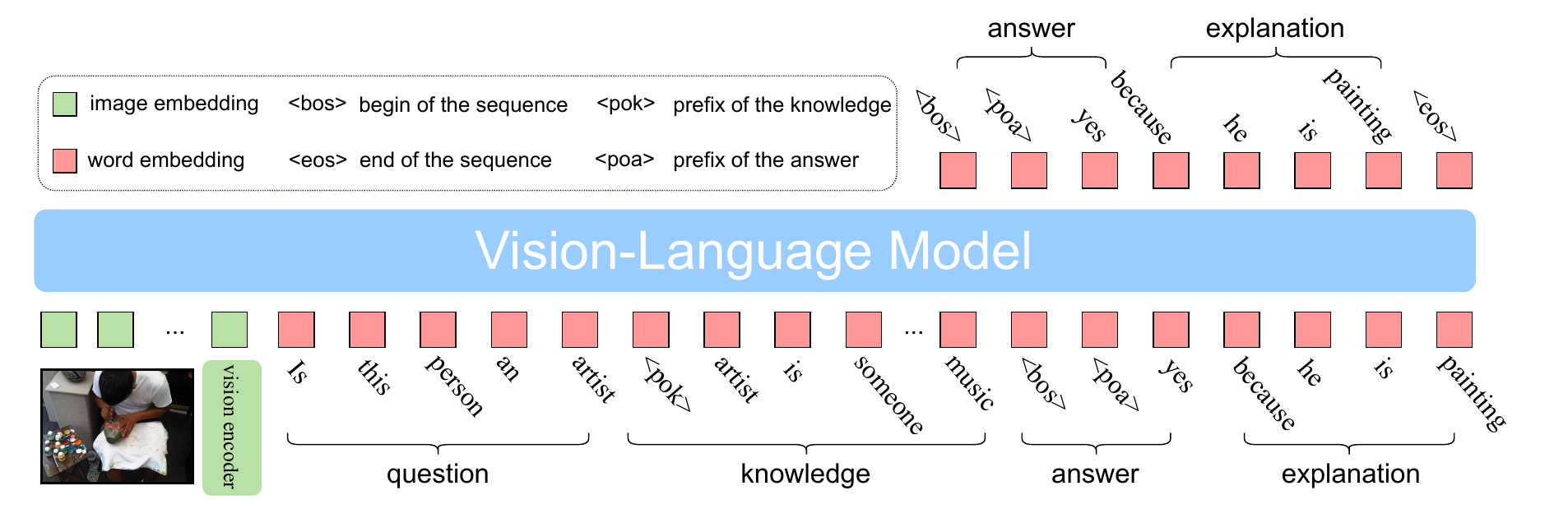}
    \caption{Architecture of our alleviation model. Given an image, its corresponding question, and external knowledge, we then append the ``<bos>'' token along with the predefined prefix for answer. The model then generates the answer and explanation in an auto-regressive manner. The external knowledge is retrieved from GPT-4o to provide additional context for the question. For the content prefix, ``<pok>'' represents the knowledge prefix (``based on the fact that'') and ``<poa>'' denotes the answer prefix (``the answer is'').}
    \label{alleviation}
\end{figure*}

\begin{figure*}[!ht]
    \centering
    \includegraphics[width=\textwidth]{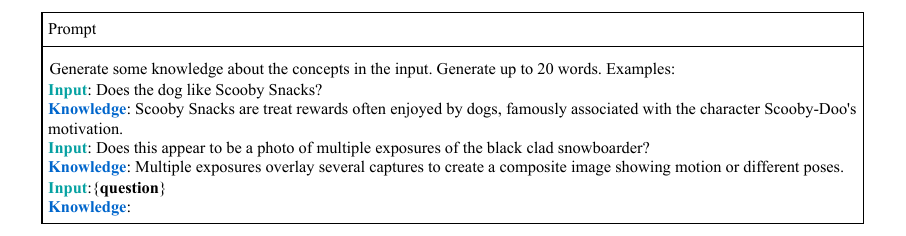}
    \caption{Prompt template for knowledge generation.}
    \label{prompt}
\end{figure*}

\subsection{Alleviation}
Because synonyms or paraphrases often cause inconsistencies between a model’s outputs and those of a benign model, a logical remedy is to ensure the model better captures synonym equivalences. To address this, we propose extending the model’s input with external knowledge relevant to the question. The key insight is that incorporating question-specific knowledge can help the model interpret synonymous words more faithfully, leading to a more robust understanding of the query. Our approach to mitigating inconsistencies in VQA-NLE involves two main steps: (1) generating question-related knowledge, and (2) injecting that knowledge into the model’s input.

\paragraph{Knowledge Generation} Drawing on \cite{questionknowledge}, we employ a heuristic method to generate question-specific knowledge. Concretely, our system prompts OpenAI’s GPT-4o API with a structured template that contains both clear instructions and illustrative examples, along with a placeholder to accommodate the new question. We then request GPT-4o to produce concise but relevant knowledge statements focusing on the core concepts of the query. Figure~\ref{prompt} outlines the complete prompt configuration, illustrating how these fixed demonstrations and guidelines steer the model toward generating succinct yet targeted knowledge.

\paragraph{Knowledge Injection}
Once the knowledge is obtained, we concatenate it with the question before feeding both into the model. Let the \textbf{image features} be
\begin{equation}
Z_I = \bigl(Z^I_1, Z^I_2, \dots, Z^I_i, \dots,Z^I_x\bigr),
\end{equation}
where each $Z^I_i \in \mathbb{R}^d$ represents the $i$-th image patch, and $x$ is the total number of image patches. For a \textbf{question} $Q = \bigl(q_1, q_2, \dots, q_t\bigr)$ (length $t$) and a \textbf{knowledge statement} $K = \bigl(k_1, k_2, \dots, k_n\bigr)$ (length $n$), we embed each word into a $d$-dimensional space via a pretrained image-caption model (DistilGPT2~\cite{nlxgpt}). This yields the \textbf{question features}
\begin{equation}
Z_Q = \bigl(Z^Q_1, Z^Q_2, \dots, Z^Q_m\bigr), 
\end{equation}
where $m$ is the total number of question tokens after tokenization, and the \textbf{knowledge features}
\begin{equation}
Z_K = \bigl(Z^K_1, Z^K_2, \dots, Z^K_r\bigr),
\end{equation}
where $r$ is the number of tokens in the knowledge statement. We then form the final \textbf{multimodal input} by concatenating all features:
\begin{equation}
\label{input_feature}
Z = \bigl(\underbrace{Z^I_1,\dots,Z^I_x}_{\text{image features}},\,
\underbrace{Z^Q_1,\dots,Z^Q_m}_{\text{question features}},\,
\underbrace{Z^K_1,\dots,Z^K_r}_{\text{knowledge features}}\bigr).
\end{equation}
Finally, we fine-tune the VQA-NLE models on this extended input until their accuracy returns to its original level. The overall architecture is shown in Figure~\ref{alleviation} where our alleviation model comprises a vision-language model and a vision encoder.
\section{Experiment}

This section offers a comprehensive assessment of our proposed attack and mitigation strategy. First, in Section~\ref{sec:setup}, we detail our experimental setup. We then present quantitative findings, examining how our attacks influence textual variations (Section~\ref{sec:results_text_based_attack}) and image manipulations (Section~\ref{sec:results_image_based_attack}). Finally, in Section~\ref{sec:case_study}, we provide a case study that offers deeper, qualitative insights into model behavior.

\subsection{Experimental Setup} \label{sec:setup}
\subsubsection{Dataset}
\noindent\textbf{VQA-X}~\cite{multimodalexplanations} \hspace{1.2ex} This vision-and-language dataset extends the original Visual Question Answering (VQA) benchmark~\cite{vqa} by appending detailed, human-written explanations to each question–answer pair. In total, VQA-X encompasses 28,000 images and 33,000 Q\&A pairs drawn from the COCO dataset~\cite{mscoco}. Of these, 29,000 pairs are allocated for training, while 1,400 are held out for validation. This additional explanatory text enables richer supervision, encouraging models to justify their answers rather than simply outputting them.

\noindent\textbf{A-OKVQA}~\cite{aokvqa} \hspace{1.2ex} A-OKVQA is another vision–language dataset that leverages images from COCO~\cite{mscoco}, but its core feature is the inclusion of rationales. Each sample thus provides a question, an answer, and a rationale explaining the reasoning behind that answer. The dataset consists of 25,000 such triplets, split into 17,100 for training and 1,100 for validation. By including explicit rationales, A-OKVQA further challenges models to exhibit both accuracy and interpretability in their responses.

\begin{table*}[!ht]
\setlength{\tabcolsep}{3.5pt}
\footnotesize
\centering
\begin{tabular}{clcccccccccccccc}
    \toprule
    & \multirow{2}{*}[-3pt]{Method} & \multicolumn{7}{c}{VQA-X} & \multicolumn{7}{c}{A-OKVQA} \\
    \cmidrule(r){3-9}\cmidrule(lr){10-16}
    & & B1 & B2 & B3 & B4 & RL & M & BS & B1 & B2 & B3 & B4 & RL & M & BS \\
    \cmidrule(lr){1-16}
    & Plural & 62.6 & 45.6 & 33.2 & 24.4 & 45.9 & 40.6 & 75.4 & 64.9 & 39.6 & 26.8 & 17.8 & 41.9 & 39.9 & 75.7 \\
    Text-based & Our Attack & \textbf{59.0} & \textbf{41.6} & \textbf{29.3} & \textbf{21.0} & \textbf{43.6} & \textbf{37.6} & \textbf{73.3} & \textbf{64.7} & \textbf{38.6} & \textbf{25.6} & \textbf{16.6} & \textbf{41.7} & \textbf{38.0} & \textbf{74.0} \\
    & Alleviation & 59.2 & 41.8 & 29.5 & 21.1 & 43.9 & 37.8 & 73.7 & 65.0 & 39.1 & 25.7 & 16.9 & 41.8 & 38.5 & 74.4 \\
    \cmidrule(lr){1-16}
    & DR & 64.1 & 46.9 & 34.5 & 25.3 & 47.0 & 41.5 & 75.7 & 63.9 & 38.9 & 26.0 & 16.9 & 41.0 & 37.9 & 74.0 \\
    Image-based & SSP & 65.3 & 48.2 & 35.4 & 25.7 & 47.1 & 42.3 & 76.0 & 65.4 & 40.4 & 27.4 & 18.2 & 42.9 & 39.6 & 74.8 \\
    & Our Attack & \textbf{59.9} & \textbf{41.6} & \textbf{29.6} & \textbf{21.2} & \textbf{43.0} & \textbf{36.9} & \textbf{73.1} & \textbf{62.0} & \textbf{37.4} & \textbf{24.3} & \textbf{15.3} & \textbf{40.4} & \textbf{36.9} & \textbf{73.8} \\
    \bottomrule
\end{tabular}
\caption{Comparison with the baselines on VQA-X and A-OKVQA datasets in the scenario of ``unfiltered'' scores.}
\label{tab:unfiltered_text_based_attack}
\end{table*}

\begin{table*}[!ht]
\setlength{\tabcolsep}{2.2pt}
\footnotesize
\centering
\begin{tabular}{clcccccccccccccccc}
    \toprule
    & \multirow{2}{*}[-3pt]{Method} & \multicolumn{8}{c}{VQA-X} & \multicolumn{8}{c}{A-OKVQA} \\
    \cmidrule(r){3-10}\cmidrule(lr){11-18}
    & & B1 & B2 & B3 & B4 & RL & M & BS & Acc & B1 & B2 & B3 & B4 & RL & M & BS & Acc \\
    \cmidrule(lr){1-18}
    & Plural & 65.9 & 49.1 & 36.5 & 27.2 & 48.5 & 43.3 & 77.1 & 74.6 & 69.8 & 45.5 & 32.3 & 23.0 & 46.4 & 45.2 & 78.5 & 37.5 \\
    Text-based & Our Attack & \textbf{63.5} & \textbf{46.5} & \textbf{33.9} & \textbf{24.8} & \textbf{46.9} & \textbf{41.2} & \textbf{75.3} & \textbf{67.2} & \textbf{69.4} & \textbf{45.4} & \textbf{31.9} & \textbf{22.4} & \textbf{46.0} & \textbf{44.0} & \textbf{77.4} & \textbf{31.7} \\
    & Alleviation & 63.8 & 46.7 & 34.0 & 24.9 & 46.9 & 41.6 & 75.7 & 68.1 & 70.7 & 46.1 & 32.2 & 22.7 & 46.6 & 44.7 & 77.5 & 30.1 \\
    \cmidrule(lr){1-18}
    & DR & 67.4 & 50.4 & 38.0 & 28.3 & 49.8 & 44.9 & 77.4 & 74.1 & 69.2 & 46.1 & 32.8 & 22.5 & 45.0 & 44.2 & 76.8 & 36.8 \\
    Image-based & SSP & 69.4 & 52.6 & 39.4 & 29.1 & 50.2 & 45.5 & 77.7 & 72.6 & 69.3 & 46.7 & 33.8 & 24.6 & 47.9 & 46.1 & 77.5 & 38.7 \\
    & Our Attack & \textbf{63.3} & \textbf{45.5} & \textbf{33.2} & \textbf{24.1} & \textbf{46.1} & \textbf{40.3} & \textbf{75.0} & \textbf{70.1} & \textbf{66.8} & \textbf{43.6} & \textbf{30.0} & \textbf{19.7} & \textbf{43.9} & \textbf{41.9} & \textbf{76.4} & \textbf{33.3} \\
    \bottomrule
\end{tabular}
\caption{Comparison with the baselines on VQA-X and A-OKVQA datasets in the scenario of ``filtered'' scores.}
\label{tab:filtered_text_based_attack}
\end{table*}

\subsubsection{Evaluation Metrics} \label{sec:evaluation_metrics}
In line with prior research~\cite{contrastivelearning}, the quality of generated explanations is measured using the following metrics: BLEU (from B1 to B4, corresponding to BLEU-1 through BLEU-4)~\cite{bleu}, METEOR (M)~\cite{meteor}, ROUGE-L (RL)~\cite{rouge}, and BERT score (BS)~\cite{bertscore}. As these datasets are for VQA tasks, we also provide accuracy to measure the correctness of the predicted answers. Additionally, we also follow~\cite{videochatgpt} to assess the VQA-NLE model for its response correctness, detail, and context comprehension.

\subsubsection{Implementation Details}
\label{sec:implementation_details}

\paragraph{Victim Models.}
Since existing works~\cite{nlxgpt, s3c, contrastivelearning} primarily rely on DistilGPT2, we adopt DistilGPT2, pretrained on image-caption pairs, as our ``victim'' model for both attacks and evaluations on the VQA-X and A-OKVQA datasets, respectively~\cite{s3c, contrastivelearning}.

\paragraph{Alleviation Model.}
To perform our knowledge-injection (alleviation) strategy, we adopt DistilGPT2~\cite{nlxgpt}, which has been pre-trained on a large corpus of image-caption pairs. We then fine-tune it separately on the VQA-X and A-OKVQA training sets, ensuring it can integrate external knowledge effectively.

\paragraph{Image Feature Extraction.}
Following prior work, we represent each image with features extracted via ViT-B/16 from CLIP~\cite{clip}. This encoder converts images into patch-level feature embeddings, which are then fed (alongside question and/or knowledge embeddings) into our models.

\paragraph{Baselines.}
We compare our attacks against adversarial attacks targeting different modalities. For image-based attacks, we adopt DR~\cite{dr} and SSP~\cite{ssp} as baselines. These approaches are designed to perturb image features exclusively and can be readily adapted to our setting. In contrast, other methods either rely entirely on classifier outputs~\cite{img_bs_n1, img_bs_n2, img_bs_n3, img_bs_n4, img_bs_n5, img_bs_n6} or combine feature perturbation with classification loss~\cite{img_bs_n7, img_bs_n8, img_bs_n9}. Such methods are incompatible with our problem setup, as pre-trained and fine-tuned models often employ different prediction heads and are optimized for distinct tasks. For text-based attacks, we refer to the method from~\cite{hypernymyinbert} as Plural, since the original work does not assign it a specific name. This approach converts singular nouns into their plural forms. To minimize semantic drift and avoid introducing contradictions, we modify only one noun per question. We adopt this method as one of our baselines for evaluating textual robustness.

\begin{table*}[!ht]
\setlength{\tabcolsep}{8pt}
\footnotesize
\centering
\begin{tabular}{cccccccc}
    \toprule
    \multirow{2}{*}[-3pt]{Method} & \multicolumn{3}{c}{VQA-X} & \multicolumn{3}{c}{A-OKVQA} \\
    \cmidrule(r){2-4}\cmidrule(lr){5-7}
     & Correctness $\downarrow$ & Detail $\downarrow$ & Context $\downarrow$ & Correctness $\downarrow$ & Detail $\downarrow$ & Context $\downarrow$ \\
    \cmidrule(lr){1-7}
    DR & 2.79 & 1.90 & 3.11 & 2.22 & 1.78 & 2.69 \\
    SSP & 2.72 & 1.88 & 3.01 & 2.26 & 1.75 & 2.77 \\
    Our Attack & \textbf{2.20} & \textbf{1.73} & \textbf{2.47} & \textbf{1.93} & \textbf{1.56} & \textbf{2.35} \\
    \bottomrule
\end{tabular}
\caption{Comparison with the baselines on VQA-X and A-OKVQA datasets in the scenario of ``unfiltered'' scores for image-based attack.}
\label{tab:unfiltered_image_based_attack}
\end{table*}

\begin{table*}[!ht]
\setlength{\tabcolsep}{6pt}
\footnotesize
\centering
\begin{tabular}{cccccccccc}
    \toprule
    \multirow{2}{*}[-3pt]{Method} & \multicolumn{4}{c}{VQA-X} & \multicolumn{4}{c}{A-OKVQA} \\
    \cmidrule(r){2-5}\cmidrule(lr){6-9}
     & Correctness $\downarrow$ & Detail $\downarrow$ & Context $\downarrow$ & Acc & Correctness $\downarrow$ & Detail $\downarrow$ & Context $\downarrow$ & Acc \\
    \cmidrule(lr){1-9}
    DR & 3.03 & 1.97 & 3.26 & 74.1 & 3.09 & 2.17 & 3.44 & 36.8 \\
    SSP & 2.98 & 1.97 & 3.19 & 72.6 & 3.00 & 2.05 & 3.36 & 38.7 \\
    Our Attack & \textbf{2.39} & \textbf{1.77} & \textbf{2.57} & \textbf{70.1} & \textbf{2.72} & \textbf{1.89} & \textbf{3.04} & \textbf{33.3} \\
    \bottomrule
\end{tabular}
\caption{Comparison with the baselines on VQA-X and A-OKVQA datasets in the scenario of ``filtered'' scores for image-based attack.}
\label{tab:filtered_image_based_attack}
\end{table*}

\subsection{Results on Text-based Attack} \label{sec:results_text_based_attack}
The results of our text-based adversarial attacks on the VQA-X and A-OKVQA datasets are detailed in Tables~\ref{tab:unfiltered_text_based_attack} and~\ref{tab:filtered_text_based_attack}. These tables differentiate between ``unfiltered'' evaluations, which assess all explanations regardless of the accuracy of the corresponding answers, and "filtered" evaluations, which consider only explanations linked to correct answers. In Table~\ref{tab:unfiltered_text_based_attack}, we illustrate that our attack not only compromises the integrity of the original model but also induces a marked reduction in the consistency of the explanations when compared to the baseline method. This effect is quantitatively substantial, with our method resulting in a 4\% decrease in BLEU-2 scores on VQA-X and a 1.9\% decrease in METEOR scores on A-OKVQA. This highlights the effectiveness of our attack in disrupting the model's ability to generate coherent and contextually appropriate explanations, thereby revealing the model's vulnerability to linguistic perturbations.

Furthermore, the incorporation of external knowledge into the model's framework has demonstrated a capability to alleviate these inconsistencies. By enhancing the contextual grounding of the explanations, this strategy not only restores but also improves their reliability, suggesting that external knowledge can serve as a countermeasure to adversarial attacks. Moving to the filtered results showcased in Table~\ref{tab:filtered_text_based_attack}, our attack methodology continues to outperform the baseline in terms of diminishing explanation consistency, thereby reinforcing the attack's effectiveness. Concurrently, our defense mechanism again proves beneficial, enhancing the consistency of explanations even when considering only correct answer contexts. This dual success underscores the comprehensive strength of our approach in both compromising and subsequently reinforcing the model's explanatory capabilities. The reduction in consistency, driven by our effective adversarial attacks, correlates strongly with a decline in accuracy, as recorded in the "Acc" column of both tables. This decline emphasizes the direct impact of our attacks on the model's overall performance, highlighting the critical link between the accuracy of answers and the coherence of explanations. These results affirm the necessity of developing more resilient models that can withstand such linguistic adversarial challenges while maintaining high standards of accuracy and explanatory depth.

\subsection{Results on Image-based Attack} \label{sec:results_image_based_attack}
We systematically modify images in a controlled fashion, expecting that the model's explanations remain consistent despite minor visual differences. By strategically removing objects that do not directly answer the question but influence the generation of explanations, we evaluate the robustness of the model in maintaining coherent explanations that align with the altered image content. Importantly, even after our proposed image edits, the attacked images maintain a high degree of similarity to their original counterparts, with average cosine similarities of 81.7\% and 82.1\% on VQA-X and A-OKVQA datasets, respectively, as measured by CLIP embeddings. This ensures that the visual changes are minimal and localized. Tables~\ref{tab:unfiltered_image_based_attack} and \ref{tab:filtered_image_based_attack} demonstrate the effectiveness of our proposed adversarial attacks, which significantly reduce the consistency of the model's responses compared to those affected by random noise. Notably, our targeted attack leads to a marked decrease in the accuracy scores on both VQA-X and A-OKVQA datasets, with the lowest recorded accuracies being 70.1\% and 33.3\%, respectively. This substantial drop indicates that the models heavily rely on the presence of specific objects to generate their answers and explanations, resulting in substantial misinterpretations of the image context when these objects are absent.

In the filtered scenario, where only correct answers contribute to the evaluation metrics, the attacked models on VQA-X and A-OKVQA exhibit notable declines in the ``Correctness,'' ``Detail,'' and ``Context'' scores compared to their performances under noise and original conditions. Specifically, ``Detail'' score reduction from 1.97 to 1.77 on VQA-X under attack conditions suggests that the removal of contextually relevant objects disrupts the model's ability to provide detailed and relevant explanations. Similarly, compared with the best baseline SSP with 3.19, our attack reduces the ``Context'' score to 2.57 on VQA-X, highlighting a deterioration in the model’s capacity to comprehend and relate to the altered visual context. This is exacerbated on the A-OKVQA dataset, where the ``Context'' score plummets to 3.04, underscoring the model's increased sensitivity to visual manipulations.

These findings underscore the vulnerability of current VQA models to targeted adversarial attacks that remove non-answering yet contextually significant objects. They also emphasize the need for developing more robust VQA systems that do not merely focus on detectable objects but comprehend the holistic scene to maintain explanation integrity under adversarial conditions.

\begin{figure*}[!ht]
    \centering
    \includegraphics[width=\textwidth]{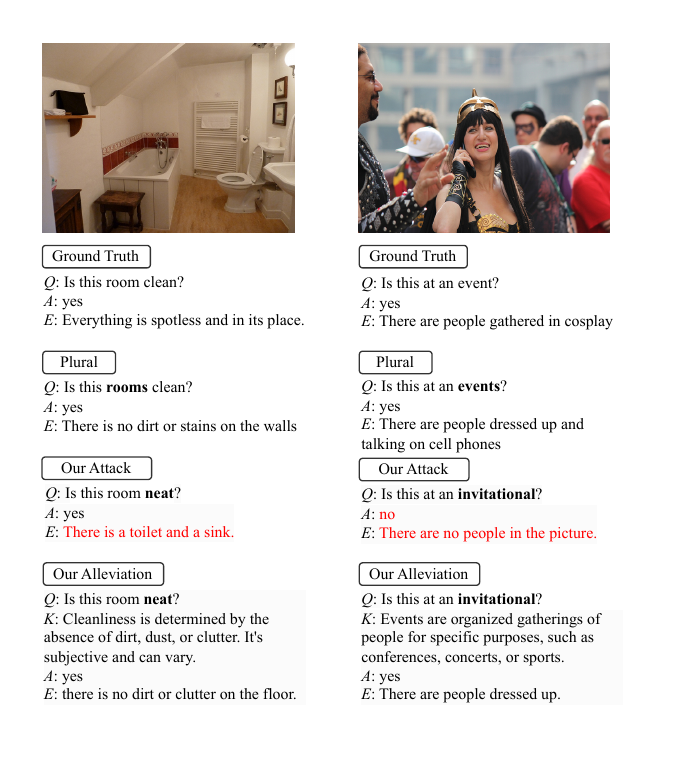}
    \vspace{-25pt}
    \caption{Qualitative examples of our text-based attack.}
    \label{text_attack_sample}
\end{figure*}

\begin{figure*}[!ht]
    \centering
    \includegraphics[width=\textwidth]{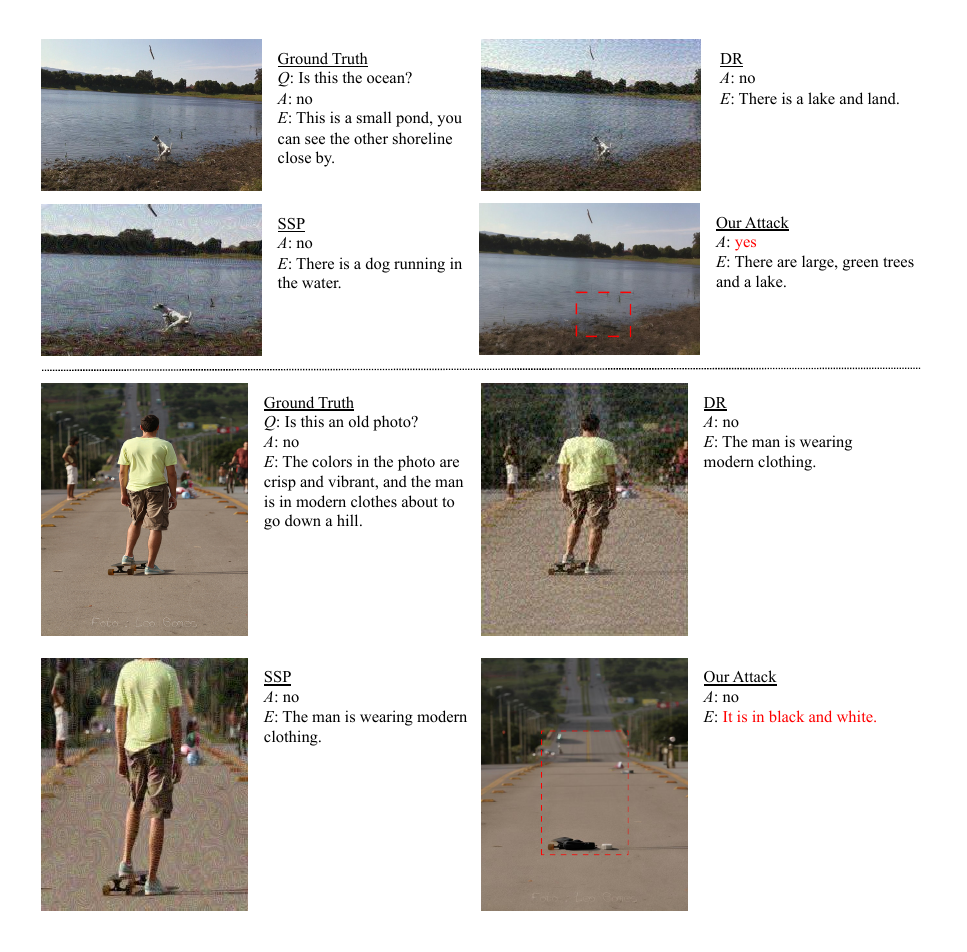}
    \vspace{-25pt}
    \caption{Qualitative examples of our image-based attack. The ``Q'' is omitted if it matches the ground truth.}
    \label{img_attack_sample}
\end{figure*}

\subsection{Case Study} \label{sec:case_study}

\paragraph{Text Attack} As illustrated in Figure~\ref{text_attack_sample} (right), the slightly changed question: \textit{``Is this at an invitational?''} yields the answer and explanation: \textit{``no because there are no people in the picture.''} This explanation contradicts the visual evidence in the image, suggesting that the model’s visual grounding has been disrupted by the subtle rephrasing. The shift leads the model to overlook relevant visual cues it had initially attended to. However, after being provided with the knowledge describing what ``event'' is, the model is able to grasp the context of the specific question, thus generating the answer and explanation that align with the question. Similarly, in Figure~\ref{text_attack_sample} (left), for the question: \textit{``Is this room neat?''}, the model correctly answers: \textit{``yes''}, but the accompanying explanation: \textit{``there is a toilet and a sink.''} is semantically misaligned. This explanation is inadequate as it merely lists objects presented in the room without addressing the notion of neatness or cleanliness, and thus fails to justify the answer. In our alleviation strategy, we guide the model by explicitly associating cleanliness with the absence of dirt. This helps steer its reasoning toward evaluating the actual tidiness of the room, resulting in a more contextually appropriate explanation.

\paragraph{Image Attack} Figure~\ref{img_attack_sample} (top) exemplifies how VQA-NLE models rely on spurious correlations rather than genuine scene understanding. Initially, the model correctly identifies it is not the ocean based on the presence of the dog in the water. However, after removing the dog from the image, the model shifts its prediction to \textit{``yes''}. This suggests that the model’s decision-making process is heavily influenced by particular objects rather than reasoning holistically about the scene. In Figure~\ref{img_attack_sample} (bottom), the model correctly explains the image is not an old photo by referring to the modern clothing worn by the man. However, after removing people from the image, the model still answers correctly with \textit{``no''}, yet the explanation becomes \textit{``it is in black and white''}, which obviously contradicts the visual evidence, suggesting a disconnect between the model’s generation and its visual grounding capabilities. Overall, the model heavily relies on superficial correlations rather than deep reasoning and contextual understanding. Instead of accurately grounding its explanations in the image and question, the model often justifies its answers using spurious associations. The VQA-NLE models struggle to adapt to minor question variations and image modifications, leading to explanations that either misalign with the question or contradict visual evidence. This indicates a fundamental gap between the model's answer generation and its ability to provide logically sound explanations.
\section{Conclusion and Future Work}

In this paper, we examine the robustness of VQA-NLE models, revealing their susceptibility to generating mutually inconsistent explanations in response to linguistic and semantic variations. To systematically evaluate these vulnerabilities, we implement BERT-Attack that perturbs input questions, and also propose a novel adversarial attack framework that modifies image content. Our experiments show that VQA-NLE models exhibit sensitivity to these perturbations, indicating a reliance on spurious correlations rather than genuine reasoning. To mitigate these inconsistencies, we introduce a method that integrates external knowledge into adversarially perturbed questions. Our results demonstrate that this approach reduces contradictions, thereby enhancing the robustness of VQA-NLE models. For future work, we plan to extend our investigation to large vision-language models such as LLaVA~\cite{llava} and Qwen-VL~\cite{qwen}. Additionally, we aim to explore the effectiveness of prompting techniques, such as chain-of-thought reasoning, as a defense mechanism against adversarial attacks by improving step-by-step reasoning.

\section*{Limitations}
Our alleviation method depends on the question-related knowledge, which may not be effective in certain cases. For example, the knowledge ``Dresses can be sleeveless or have varying sleeve styles, such as short, long, or cap sleeves.'' extracted from the benign question ``Does the dress have sleeves?'' is helpful for the adversarial question ``Does the gown have sleeves?'' because it relates ``gown'' with ``dress'' while guiding models to focus on sleeves. Meanwhile, ``The dress could refer to a specific dress that gained viral attention in 2015 due to the optical illusion of its colors.'' provides little relevant information for answering the question, making it a poor knowledge statement.
\bibliography{custom}

\appendix

\end{document}